\documentclass[letterpaper, 10 pt, conference]{ieeeconf}  

\IEEEoverridecommandlockouts                              

\usepackage{graphics} 
\usepackage{graphicx}

\usepackage{epsfig} 
\usepackage{amsmath} 
\usepackage{amssymb}  

\usepackage{algorithm}
\usepackage{algpseudocode}
\algnewcommand\AAND{\textbf{ and }}
\algnewcommand\Or{\textbf{ or }}

\usepackage{color}
\usepackage{citesort}
\usepackage{url}
\usepackage{makecell}

\DeclareMathAlphabet{\pazocal}{OMS}{zplm}{m}{n}

\newcommand{\Bs}{\pazocal{B}}

\DeclareMathAlphabet{\mathpzc}{OT1}{pzc}{m}{it}

\usepackage{array}
\newcolumntype{C}[1]{>{\centering\arraybackslash}p{#1}}
\newcolumntype{M}[1]{>{\raggedright\arraybackslash}p{#1}}

\usepackage{array} 
\newcolumntype{L}[1]{>{\raggedright\let\newline\\\arraybackslash\hspace{0pt}}m{#1}}	
\newcolumntype{S}[1]{>{\centering\let\newline\\\arraybackslash\hspace{0pt}}m{#1}}
\newcolumntype{R}[1]{>{\raggedleft\let\newline\\\arraybackslash\hspace{0pt}}m{#1}}

\makeatletter
\renewcommand*{\@opargbegintheorem}[3]{\trivlist
  \item[\hskip \labelsep{\itshape #1\ #2}] \textit{(#3)}\ }
\makeatother

\title{\LARGE \bf
Motion Primitives-based Navigation Planning\\ using Deep Collision Prediction
}

\author{Huan Nguyen, Sondre Holm Fyhn, Paolo De Petris, and Kostas Alexis
\thanks{This material was supported by Research Council of Norway under project SENTIENT (grant number 321435).}
\thanks{The authors are with the Autonomous Robots Lab, Norwegian University of Science and Technology (NTNU), O. S. Bragstads Plass 2D, 7034, Trondheim, Norway {\tt\small dinh.h.nguyen@ntnu.no}}
}

\begin{document}

\maketitle
\thispagestyle{empty}
\pagestyle{empty}

\begin{abstract}

This paper contributes a method to design a novel navigation planner exploiting a learning-based collision prediction network. The neural network is tasked to predict the collision cost of each action sequence in a predefined motion primitives library in the robot's velocity-steering angle space, given only the current depth image and the estimated linear and angular velocities of the robot. Furthermore, we account for the uncertainty of the robot's partial state by utilizing the Unscented Transform and the uncertainty of the neural network model by using Monte Carlo dropout. The uncertainty-aware collision cost is then combined with the goal direction given by a global planner in order to determine the best action sequence to execute in a receding horizon manner. To demonstrate the method, we develop a resilient small flying robot integrating lightweight sensing and computing resources. A set of simulation and experimental studies, including a field deployment, in both cluttered and perceptually-challenging environments is conducted to evaluate the quality of the prediction network and the performance of the proposed planner.

\end{abstract}

\section{INTRODUCTION}\label{sec:intro}

Recent advances in the field of aerial robotics have enabled their utilization in various applications including industrial inspection~\cite{BABOOMS_ICRA_15,SIP_AURO_2015}, search and rescue~\cite{tomic2012toward}, surveillance~\cite{grocholsky2006cooperative}, subterranean exploration~\cite{GBPLANNER_JFR_2020}, and more. Despite the progress, the task of autonomous navigation in complex, perceptually-degraded environments, such as obstacle-filled industrial settings, and underground mines remains particularly challenging. In such settings, the underlying robot localization and mapping may be subject to significant uncertainty and drift~\cite{Shehryar2020CompSLAM,Zhao2020TPTIO,Palieri2021LOCUS}, and requires high computational cost~\cite{Han2019FIESTA}, while consistent high-resolution mapping can be particularly demanding. In this context, ensuring collision-free navigation can be a gruelling task and remains an open problem.

%
\begin{figure}[h!]
\centering
    \includegraphics[width=0.99\columnwidth]{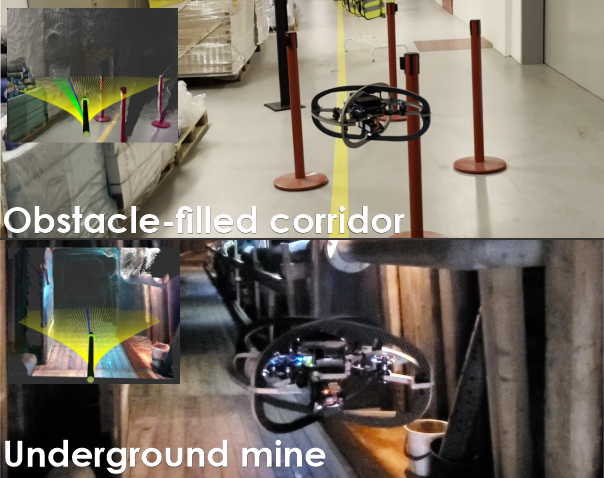}
\vspace{-5ex}
\caption{Instances of experimental evaluation of the proposed motion primitives-based planner using a deep collision predictor.}
\vspace{-4ex}
\label{fig:oracleintro}
\end{figure}
%

In this work, we aim to take a different approach to allow safe navigation of aerial robots by departing from the classical techniques that require a continuously maintained and updated map of the environment within which the robot localizes~\cite{burri2015real}. Employing a data-driven approach to allow collision-free navigation without explicit map building and an accurate estimate of the robot's position, we propose a local mOtion pRimitives-bAsed navigation planner using a deep CoLlision prEdictor (ORACLE) that further accounts for the goal direction provided by a global planner. The collision predictor is a deep neural network taking as inputs the current depth image, estimated linear and angular velocities and associated covariance, as well as action sequences from a fixed motion primitives library based on forward velocity and steering angle commands. It evaluates - in parallel - the collision probability for each action sequence, based on which the safest motion primitives are chosen and the corresponding goal-reaching costs are calculated to determine the best action in a receding horizon fashion.

The proposed approach was experimentally evaluated in two unseen real-life environments, while the complete method was designed and trained purely in simulation. As a testbed for our learning-based navigation policy, we developed a new collision-tolerant aerial robot, called Learning-based Micro Flyer (LMF). LMF integrates a lightweight, and low-cost sensor suite including an Intel Realsense D455 RGB-D sensor, an Intel Realsense T265 visual-inertial module, and a $1\textrm{D}$ range sensor, as well as an autopilot, and an NVIDIA Xavier NX CPU/GPU board. The first experiment was conducted within obstacle-filled building corridors and aimed to verify the ability of the method to avoid novel obstacles, while navigating using goal directions commanded by a global planner. The second experiment was carried out in an underground mine environment near Trondheim, Norway, demonstrating the ability of the robot to safely navigate in a narrow, relatively dark environment that can lead to high uncertainty in visual-inertial estimation.

The remainder of this paper is organized as follows: Section~\ref{sec:related} presents related work, followed by the problem statement in Section~\ref{sec:probstat}. The proposed learning-based navigation policy is presented in Section~\ref{sec:approach}. Evaluation studies are detailed in Section~\ref{sec:evaluation}, followed by conclusions in Section~\ref{sec:concl}.

\section{RELATED WORK}\label{sec:related}
A set of contributions in a) planning with neural networks, b) planning under uncertainty and modeling uncertainty in deep neural networks, and c) planning with motion primitives relate to this work. 

\subsection{Planning with Neural Networks}
A set of works have applied deep learning to the problem of autonomous robot navigation. Specifically, the problem of navigation with RGB and depth cameras has attracted increased attention as these sensors are low-cost, low-power and lightweight. The authors in~\cite{Gandhi2017learning2fly,Loquercio2018Dronet} use supervised learning, while the work in~\cite{Fereshteh2017CAD2RL} utilizes reinforcement learning and domain randomization with RGB image inputs to learn reactive navigation policies.
While the robot's state is often available for the low-level control, these works do not utilize the current estimated state of the robot to make predictions although this can lead to better performance. On the other hand, the authors in~\cite{Tolani2021visual} use the current robot's linear and angular velocities in addition to the latest camera frame to generate position setpoints. However, position feedback is not always available or reliable, especially in perceptually degraded environments~\cite{Shehryar2020CompSLAM}. Neural networks are used in~\cite{Kahn2021BADGR} and~\cite{Kahn2021LAND} to predict probabilities of events such as collisions, going over bumpy terrain, or human disengagement. Nevertheless, their action library does not explicitly account for the limited Field Of View (FOV) of RGB sensors, and combining goal-reaching and collision-avoidance rewards into one reward function may not lead to safe behaviour
when the goal-reaching reward outweighs the collision-avoidance reward. 
Other works utilize the depth image as an input, which is also employed in our work, either by using the latest data~\cite{Tai2018GAIL}, or a sliding window of depth images~\cite{Hoeller2021representation}. However, these works do not account for the uncertainty in the estimated state of the robot and the neural network model.

\subsection{Planning under Uncertainty and Modeling Uncertainty in Deep Learning}
Modeling uncertainty is essential to achieve safe planning in environments where the state of the robot is highly uncertain~\cite{bry2011rapidly}. Moreover, when using deep neural networks for making predictions, there are two kinds of uncertainty that need to be considered: aleatoric uncertainty which captures noise inherent in the observations, and epistemic uncertainty which accounts for model uncertainty~\cite{Kendall2017uncertainty}. Most existing works applying deep neural networks for autonomous navigation account for epistemic uncertainty only, for instance by using autoencoders~\cite{Richter2017SafeVN}, dropout and bootstrap~\cite{Kahn2017uncertainty,Lutjens2019safeRL}. One of the exceptions is~\cite{Loquercio2020uncertainty} which accounts for both uncertainty in the sensor observations and model uncertainty. However, the method for propagating aleatoric uncertainty through sigmoid and tanh activation layers, which are usually used in recurrent neural networks, is not presented. In this work, we approximately propagate the robot's state uncertainty through the prediction network using the Unscented Transform (UT)~\cite{Julier1997unscented} as in~\cite{Abdelaziz2015propagate}, and use Monte Carlo (MC) dropout~\cite{Gal2016dropout,gal2016bayesian} to account for the model uncertainty.

\subsection{Planning with Motion Primitives}
A motion primitives library~\cite{bottasso2008path} can be generated by either sampling the vehicle's configuration space or its control space~\cite{Howard2008sampling}. 
Depending on what type of low-level controller is implemented, control space primitives for multirotors can be generated by sampling jerk~\cite{Liu2017searchbased}, acceleration~\cite{MBPlanner_ICRA_2019,Florence2020integrated}, or other input signals. The authors in~\cite{Bucki2020rappids} use configuration space primitives proposed in~\cite{Mueller2015motionprimitive} to generate minimum-jerk primitives online, however, position feedback is required and not all generated trajectories are guaranteed to lie within the sensor's FOV. Our work follows approaches like those  proposed in~\cite{Lopez2017aggressive,goel2021fastexploration} to sample the desired speed and steering angle, where possible steering angles are uniformly sampled from within the FOV.


\section{PROBLEM FORMULATION}\label{sec:probstat}
The broader problem considered in this work is that of autonomous collision-free aerial robot navigation assuming no access to the maps of the environment (from offline data or online reconstruction) and no information for the robot position but only a partial state estimate of the robot's orientation and linear/angular velocities combined with real-time depth data within an angle- and range-constrained sensor frustum. We assume that there is a global planner providing goal directions to the robot (e.g., for exploration), possibly by having access to a topological map of the environment, and focus on designing a local safe navigation planner to avoid obstacles while following the goal directions. Let $\Bs$ be the body frame of the robot, $o_t$ be the current depth image, $s_t~(s_t=[v_t, \omega_t]^T)$ the estimated partial state of the robot including the forward velocity in $\Bs~(v_t)$ and the angular velocity around z-axis of $\Bs~(\omega_t)$, $\Sigma_t=\text{diag}([\sigma_{v,t}^2, \Sigma_{\omega,t}^2])$ the covariance of the estimated robot's partial state, $\psi^g_t$ the goal heading direction given by the global planner, and $a_{t:t+H}=[a_t, a_{t+1},\hdots,a_{t+H-1}]$ an action sequence having length $H$ where the action at time step $t+i~(i=0,\hdots,H-1)$ includes the reference forward speed $v^r_{t+i}$ and the steering angle from the current yaw angle of the robot $\psi^r_{t+i}~(a_{t+i}=[v^r_{t+i}, \psi^r_{t+i}]^T)$. The exact problem considered is then formulated as that of finding an optimized collision-free sequence of actions $a_{t:t+H}$ enabling the robot to safely move towards the goal direction $\psi^g_t$ given $(o_t,s_t,\Sigma_t)$. 

\section{PROPOSED APPROACH}\label{sec:approach}
%
\begin{figure*}[h!]
\centering
    \includegraphics[width=0.92\textwidth]{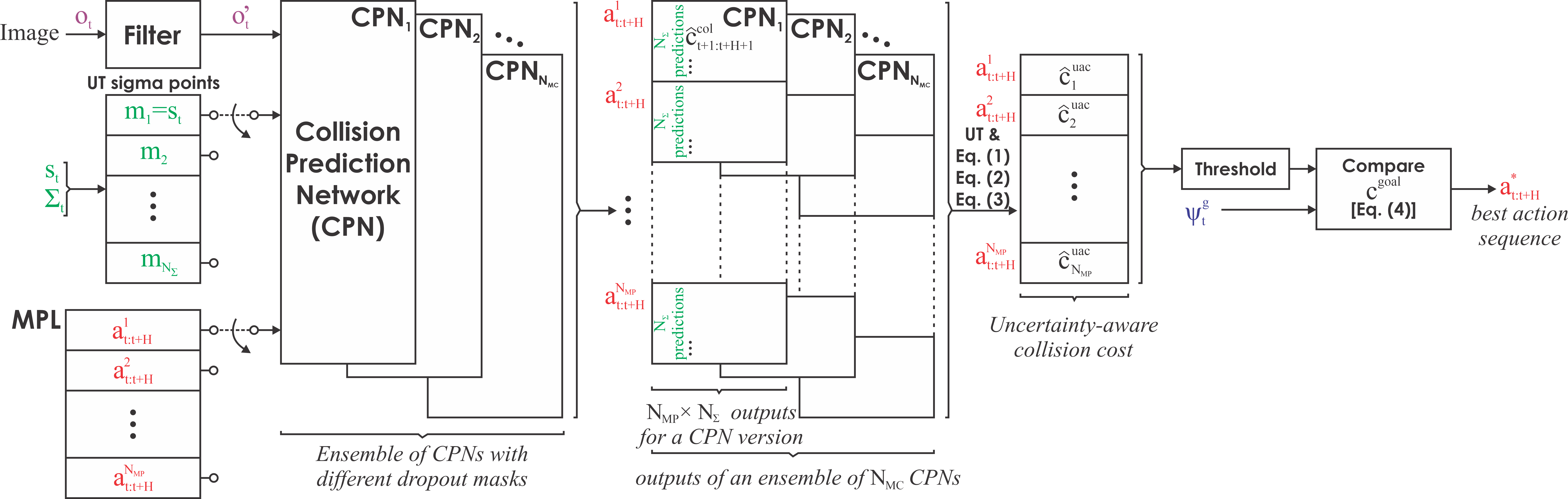}
\vspace{-2ex}
\caption{Overview of the algorithmic architecture of ORACLE.}\label{fig:ORACLEoverview}
\vspace{-3ex}
\end{figure*}
%

The proposed mOtion pRimitives-bAsed navigation planner using a deep CoLlision prEdictor (ORACLE) is detailed in this section. 
Assuming only access to estimates of the robot's linear and angular velocities, alongside the depth image, ORACLE identifies the next best sequence of actions (velocity-steering commands) that ensure that the system is navigating towards a goal heading direction, while avoiding the obstacles in its environment. The first action of this sequence is executed by the robot, while the process continues iteratively. Importantly, the ``global'' goal direction may be provided by any global planner thus allowing ORACLE to be combined with any high level planning framework. Figure~\ref{fig:ORACLEoverview} provides an overview of the method's architecture.

\subsection{Velocity-Steering Angle Motion Primitives Library}
In this work, each candidate action sequence is a time sequence of velocity and steering commands derived from a predefined Motion Primitives Library (MPL) and has fixed forward speed and fixed steering angle sampled within the FOV of the depth sensor for every time step $v^r_{t+i}=v^r,\psi^r_{t+i}=\psi^r~(i=0,\hdots,H-1)$. Specifically, we denote $a^k_{t:t+H}$ as the $k^\text{th}$ action sequence in the MPL. Each planned sequence is such that it guarantees that the trajectories always lie in the FOV of the sensor. As a notable distinction to other MPL-based methods, the planned sequences do not include the robot's position space but remain in velocity/steering angle space as the underlying assumption is that ORACLE does not have access to a position estimate. 




\subsection{Learning-based Collision Prediction Network (CPN)}

At the core of ORACLE is a neural network that processes a) the input depth image $o_t$, b) the robot's partial state $s_t$, and c) motion primitives-based sequences of future velocity and steering angle references $a_{t:t+H}$, and is trained to predict the collision probabilities of the anticipated robot motion at each time step from $t+1$ to $t+H$ in the future $\hat{c}^{col}_{t+1:t+H+1}=[\hat{c}^{col}_{t+1}, \hat{c}^{col}_{t+2},\hdots,\hat{c}^{col}_{t+H}]$ by entirely using collision data in simulation. Thus, ORACLE avoids the need for hand-engineered collision checking algorithms such as~\cite{Bucki2020rappids,Gao2019FlyingOnPCL} or access to a reconstructed map of the environment~\cite{hornung13auro,voxblox}. 

Since real-life depth images are often subject to several shortcomings compared to simulated data, including a) missing information, b) loss of detail, and c) depth noise~\cite{Hoeller2021representation}, we perform an additional filtering step using the IP-Basic algorithm~\cite{Ku2018InDO} to refine the depth frame and thus reduce the mismatch between the real and simulated depth images. The collision costs for every action sequence in the motion primitives library of velocity-steering commands can then be evaluated in parallel as per~\cite{kew2019neuralcollision} exploiting modern GPU architectures and thus enabling high update rate compute. Notably, when evaluating the collision costs, ORACLE does not only consider the mean estimate of the robot's partial state but also the estimated uncertainty, as well as the uncertainty in the neural network model. 

\begin{figure}[h!]
\centering
    \includegraphics[width=0.99\columnwidth]{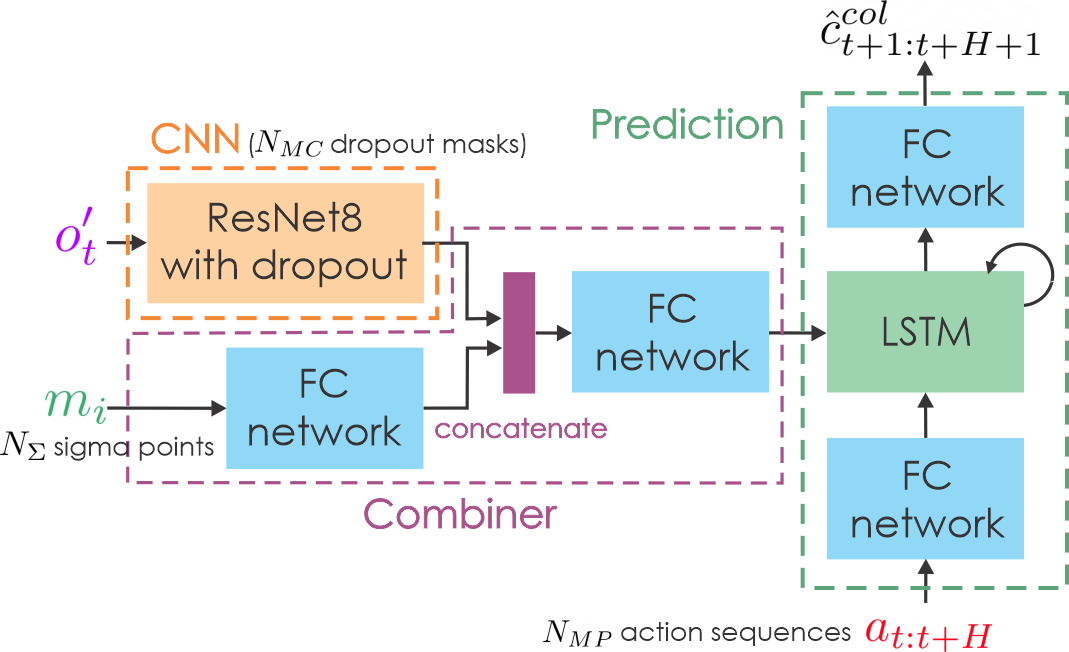}
\vspace{-3ex}
\caption{Architecture of the Collision Prediction Network (CPN).} \label{fig:NN_architecture}
\vspace{-3ex}
\end{figure}

In further detail, the filtered depth image $o_t^\prime$ of size $270\times 480$ is processed by a Convolutional Neural Network (CNN) branch, while the estimated partial state of the robot is processed by a Fully-Connected (FC) network before concatenating it with the output feature vector from the CNN branch. The concatenated vector is then fed to another FC network before being used as the initial hidden state of a Long Short-Term Memory (LSTM) network. The input to the LSTM cells is generated by the velocity-steering angle action sequence provided by the motion primitives library, while the outputs of the LSTM cells are passed through a FC network to predict the collision label at each time step. The prediction network architecture, shown in Figure~\ref{fig:NN_architecture}, is inspired by the network in~\cite{Kahn2021LAND}. However, we replace the MobileNetV2 part with the ResNet8 network with dropout layers as in~\cite{Loquercio2018Dronet} for faster inference speed and model uncertainty estimation using MC dropout as in~\cite{gal2016bayesian}. For the type of training considered, typically there are more negative (non-collision) labels than positive (collision) labels in the dataset. Hence, we use a higher weight for positive labels to ensure high recall which leads to a safer behaviour. The network is trained end-to-end with weighted binary cross-entropy loss using the Adam optimizer.

%
%

\subsection{Uncertainty-aware Prediction}\label{sec:uncertainty}

As mentioned, ORACLE further considers the uncertainty of the robot's partial state and the model uncertainty of the collision prediction network. First, we calculate the final collision cost for each action sequence in the motion primitives library as the weighted sum of the collision probabilities at each time step. Specifically, the sooner the collision event is predicted to happen, the more it will contribute to the final collision cost:

\begin{equation} \label{eq:col_cost}
    \hat{c}^{col} = \sum_{i=1}^{H}{\hat{c}^{col}_{t+i} e^{-\lambda (i-1)}},~\lambda > 0
\end{equation} 
To account for the uncertainty of $s_t$, which cannot be negligible - especially in fast flight or within perceptually degraded environments - we utilize the Unscented Transform (UT)~\cite{Julier1997unscented} to approximately propagate the uncertainty in $s_t$ to the predicted collision cost $\hat{c}^{col}$ of an action sequence $a_{t:t+H}$. In the UT, for a $k$-dimensional robot state, $N_{\Sigma}=2k + 1$ sigma points, and their associated weights, are computed based on the mean value $s_t$ and covariance $\Sigma_t$. These sigma points are then propagated though the CPN, and the mean and variance of the output distribution of $\hat{c}^{col}$ are calculated based on the output predictions of the sigma points.

Additionally, the model uncertainty - which can be significant in novel environments - can be captured by performing MC dropout for the network as in~\cite{gal2016bayesian}. For efficient neural network forward pass and uncertainty estimation, we only perform MC dropout in the CNN branch and split the neural network shown in Figure~\ref{fig:NN_architecture} into $3$ parts, namely the CNN, Combiner, and Prediction networks. Let a) $N_{MC}, N_{MP}$ be the number of dropout masks and action sequences in the MPL, respectively, and b) $\sigma^{col}_{n}, \mu^{col}_{n}~(n=1,\hdots,N_{MC})$ the variance and mean of the predicted collision cost of $a_{t:t+H}$, estimated by the UT with different dropout masks of the CNN part. The total variance can then be expressed as:

\begin{equation} \label{eq:total_variance}
    \sigma^{col}_{tot} = \frac{1}{N_{MC}} \sum_{n=1}^{N_{MC}}{\left[\sigma^{col}_{n} + (\mu^{col}_{n} - \Bar{\mu}^{col})^2\right]}
\end{equation}
where $\Bar{\mu}^{col} = \frac{1}{N_{MC}}\sum_{n=1}^{N_{MC}}{\mu^{col}_{n}}$. The final uncertainty-aware collision cost for an action sequence is given as:

\begin{equation} \label{eq:uncertain_col_cost}
    \hat{c}^{uac} = \Bar{\mu}^{col} + \alpha \sqrt{\sigma^{col}_{tot}},~\alpha > 0
\end{equation}
Specifically, we denote $\hat{c}_k^{uac}$ as the uncertainty-aware collision cost of the $k^\text{th}$ action sequence $a^k_{t:t+H}$ in the MPL ($k=1,\hdots,N_{MP}$). By splitting the prediction network into $3$ parts, we can perform inference on the ensemble of CNNs, the Combiner and Prediction networks with different batch sizes of $1,~N_{MC}\times N_{\Sigma}$, and $N_{MC} \times N_{\Sigma} \times N_{MP}$, respectively, avoiding the need to run the forward passes through the whole CPN $N_{MC} \times N_{\Sigma} \times N_{MP}$ times.

\subsection{Learning-based Navigation Policy}

After calculating the uncertainty-aware predicted collision cost for each action sequence in the motion primitives library, as described in section~\ref{sec:uncertainty}, the minimum collision cost $\hat{c}^{uac}_{min}$ of all action sequences is calculated, and all the action sequences having collision cost greater than $\hat{c}^{uac}_{min} + c_{th}$, where $c_{th}$ is a set positive threshold, are discarded. The remaining action sequences are checked for deviation from the global planner's goal direction $\psi^g_t$:

\begin{equation} \label{eq:goal_cost}
    c^{goal} = |\text{wrap}(\psi^r + \psi_t - \psi^g_t)|
\end{equation}
where $\psi_t$ is the current yaw angle of the robot and $\text{wrap}(.)$ is the function that wraps an angle in radians to $[-\pi,\pi]$. The best action sequence is chosen and its first step is executed, while the whole procedure is repeated in a receding horizon fashion. Algorithm~\ref{alg:planning_process} outlines ORACLE's key steps.

\begin{algorithm}[h]
\caption{ORACLE Navigation Planner} \label{alg:planning_process}
\begin{algorithmic}[1]
\State \emph{$s_t, \Sigma_t \gets$ Get robot's partial state and covariance}
\State \emph{$m_i~(i=1,\hdots,N_{\Sigma}) \gets$ sigma points given by UT} 
\State \emph{$o^\prime_t \gets$ Get filtered depth image}
\For{$n=1$ to $N_{MC}$}
    \State \emph{$CPN_{n} \gets$ Apply dropout mask $n$ to $CPN$}
    \For{$a_{t:t+H}$ in $MPL$}
        \For{$i=1$ to $N_{\Sigma}$}
            \State \emph{// Parallel evaluation in GPU}
            \State \emph{$\hat{c}^{col}_{t+1:t+H+1} \gets CPN_n(o^\prime_t,m_i,a_{t:t+H})$}
            \State \emph{Calculate $\hat{c}^{col}$ by Eq.~(\ref{eq:col_cost})}
        \EndFor
        \State \emph{Calculate $\sigma^{col}_{n}, \mu^{col}_{n}$ by UT on $\{\hat{c}^{col}\}_{N_{\Sigma}\times 1}$}
    \EndFor
\EndFor
\For{$a_{t:t+H}$ in $MPL$}
    \State \emph{Calculate $\hat{c}^{uac}$ by Eq.~(\ref{eq:total_variance}),~(\ref{eq:uncertain_col_cost})}
\EndFor
\State \emph{$\hat{c}^{uac}_{min} \gets \min_{k=1,...,N_{MP}} \{ \hat{c}^{uac}_k \} $}
\For{$a_{t:t+H}$ in $MPL$ having $\hat{c}^{uac} < \hat{c}^{uac}_{min} + c_{th}$}
    \State \emph{Calculate ${c}^{goal}$ by Eq.~(\ref{eq:goal_cost})}
\EndFor
\State \emph{Choose $a^{*}_{t:t+H}$ in $MPL$ having the smallest ${c}^{goal}$}
\State \emph{Execute the first action $a^{*}_t$ then go to step $1$}
\end{algorithmic}
\end{algorithm}



\subsection{Implementation Details and Data Collection} \label{subsec:implementation}

The RotorS simulator~\cite{furrer2016rotors} is used to collect data for training the collision predictor network. To ensure successful sim-to-real transfer, the dynamics of the low-level attitude and velocity controllers of the quadrotor model used in the simulator are tuned to match the real system. Relevant methods for dynamic system identification of MAVs are presented in~\cite{sa2017sysid}. To collect data for predicting collision probabilities at the future time steps, an action sequence with random $v^r$ and $\psi^r$ within the sensor's FOV is drawn and is fully executed. This process is repeated until the robot collides with the obstacles or a timeout event occurs. One training data point $d$ is recorded every time the robot moves more than $\Delta_{th}$ meters, having the format $d=(o_t, s_t, a_{t:t+H}, c^{col}_{t+1:t+H+1})$ where $c^{col}_{t+1:t+H+1}=[c^{col}_{t+1},c^{col}_{t+2},\hdots,c^{col}_{t+H}],~c^{col}_{t+i}$ denotes the collision label at time step $t+i,~i=1,\hdots,H$ (equal to $1$ for collision and $0$ for non-collision status). When the collision happens midway an action sequence, for instance after the execution of $a_{t+k}~(k\leq H)$, then the collision labels of the remaining time steps $c^{col}_{t+k+1:t+H}$ are set to $1$, and augmented data points are also added to the dataset by replacing the actions after $a_{t+k}$ with randomly sampled actions as in~\cite{Kahn2021LAND}. The number of data points created by augmenting the remaining actions is such that the number of data points with no collision label and the number of data points with at least one collision label are almost equal. Moreover, given that the robot dynamics are holonomic and effectively invariant to its heading orientation, we also perform horizontal flip data augmentation, specifically by adding the augmented data point $d^{flip}=(o^{flip}_t, s^{flip}_t, a^{flip}_{t:t+H}, c^{col}_{t+1:t+H+1})$, where $o^{flip}_t$ is the horizontally flipped image of $o_t$, $s^{flip}_t$ and $a^{flip}_{t:t+H}$ are created by changing the sign of $\omega_{t}$ and $\psi^r_{t+i}~(i=0,\hdots,H-1)$ in $s_{t}$ and $a_{t:t+H}$, respectively. 

\begin{figure}[h!]
\centering
    \includegraphics[width=0.99\columnwidth]{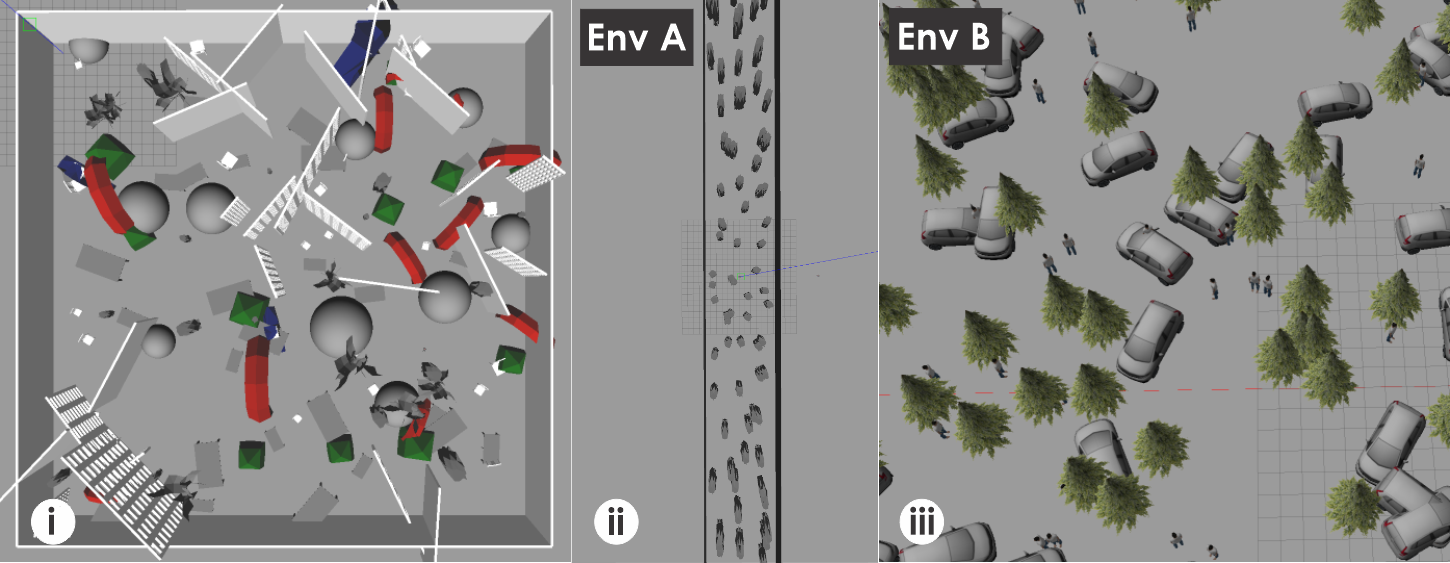}
\vspace{-3ex}
\caption{i) Indicative simulation environments for collecting training data, ii,iii) environments for simulation studies evaluating ORACLE.}\label{fig:sim_envs}
\vspace{-2ex}
\end{figure}

In order to collect a comprehensive dataset for training the collision predictor, we randomized the initial position and orientation of the robot, as well as the obstacles' locations, categories, dimensions, and densities in order to collect around 1.5 million data points in total, Figure~\ref{fig:sim_envs}.i illustrates one indicative training environment. The derived dataset was then split into a training and validation subset. Notably, a particular limitation of the depth sensor used in real-life was taken into account. Specifically, to tackle the problem that depth cameras often present gaps in depth data when facing a reflective surface~\cite{Popovic2021DepthCompletion}, we designed artificial obstacles having holes with sizes that could ``just'' fit our flying robot, or smaller, to enhance the trained network's tendency to avoid such extremely narrow or erroneously detected ``passages''. After training, the CPN achieves a prediction accuracy of $98.15\%$, precision of $98.3\%$ and recall of $97.4\%$ on the validation dataset. A page with further details is maintained at \url{https://s.ntnu.no/oracle}.

\section{EVALUATION STUDIES}\label{sec:evaluation}
A set of evaluation studies were then conducted to verify the proposed learning-based navigation planner. 


%
%

\subsection{Simulation Studies}
To evaluate ORACLE's ability to negotiate novel environments in combination with highly degraded state estimates, we conducted simulation studies and compare it with a more ``naive'' version which utilizes the CPN directly to calculate the collision cost without considering the uncertainty of $s_t$ or that of the neural network model as in~\cite{Kahn2021BADGR}. Accordingly, the naive planner is not using the UT samples and the ensemble of NNs (cf. Figure~\ref{fig:ORACLEoverview}).
Two environments are created to verify the method's performance, namely a) environment A (Figure~\ref{fig:sim_envs}.ii) that contains obstacles seen during the training phase and the robot is given perfect state estimation, while b) environment B (Figure~\ref{fig:sim_envs}.iii) contains unseen obstacles and the robot is given largely deteriorated state estimation $s_t^{det} = [v^{det}_t, ~\omega^{det}_t]^T$ (specifically $v^{det}_t=v_t - 1.0 \text{m/s}, ~\omega^{det}_t=\omega_t + 0.1 \text{rad/s}$). Both ORACLE and the naive planner are deployed in each environment $20$ times with different initial conditions of the robot, a timeout period of $100$ seconds, and reference forward velocity $1.25\textrm{m/s}$. The total number of collisions, and the average flight time before a collision are reported in Table~\ref{tab:sim_result}. Both planners perform well in environment A with the average flight times almost equal to the timeout period. However, the uncertainty-aware ORACLE significantly outperforms the naive planner in environment B where uncertainty-aware collision prediction is essential.    

\begin{table}[h!]
 \vspace{-2ex}
\caption{Simulation results.} \label{tab:sim_result}
\centering
\begin{tabular}{c|ccc}  
\hline
\textbf{Environment} & \textbf{Planner} & \textbf{Collisions} & \makecell{\textbf{Average} \\ \textbf{flight time} } \\
\hline
     A: Seen obstacles         & Naive          & 0/20  & 100s \\
    + perfect state         & Uncertainty-aware & 0/20  & 100s \\
\hline
     B: Novel obstacles        & Naive          & 20/20 & 25.9s \\
    + deteriorated state    & Uncertainty-aware & 10/20 & 62.9s \\            
\hline
\end{tabular}
 \vspace{-2ex}
\end{table}

\subsection{Experimental Studies}

Two real-life experiments were conducted using the LMF platform to evaluate the performance of ORACLE.

%
\begin{figure}[h!]
\centering
    \includegraphics[width=0.99\columnwidth]{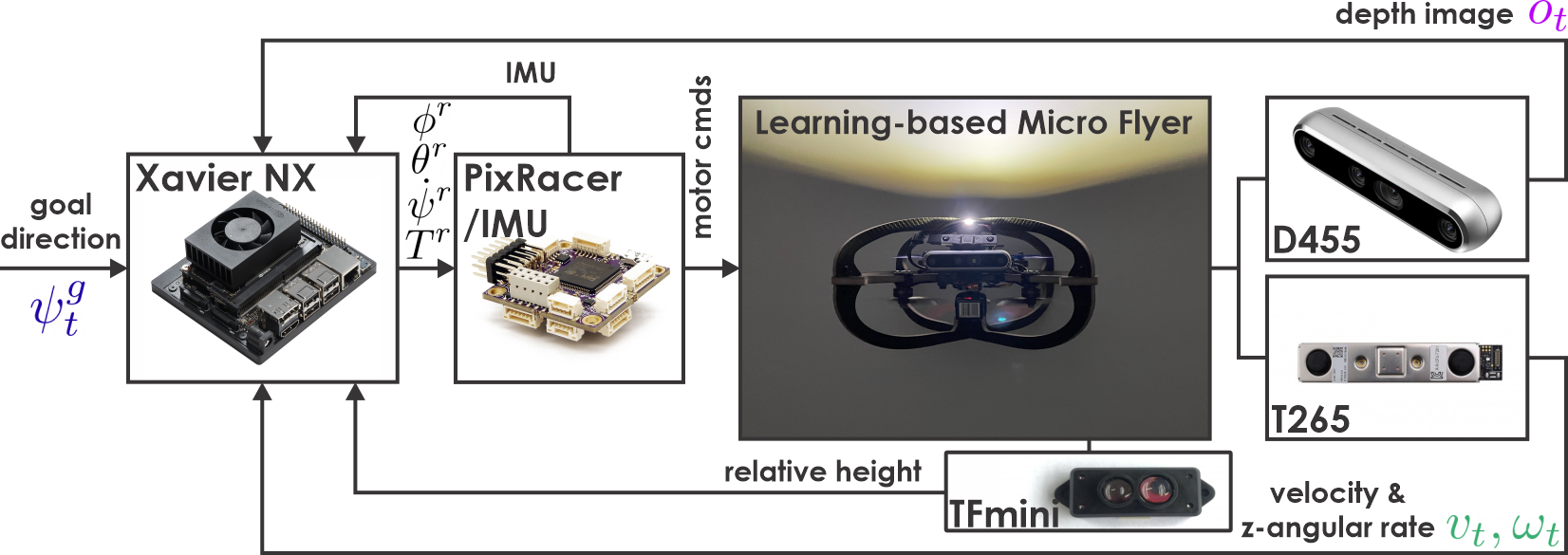}
\vspace{-3ex}
\caption{Main hardware components onboard LMF.}\label{fig:radmfmotivation}
\vspace{-2ex}
\end{figure}
%

%
\begin{figure*}[h!]
\centering
    \includegraphics[width=0.99\textwidth]{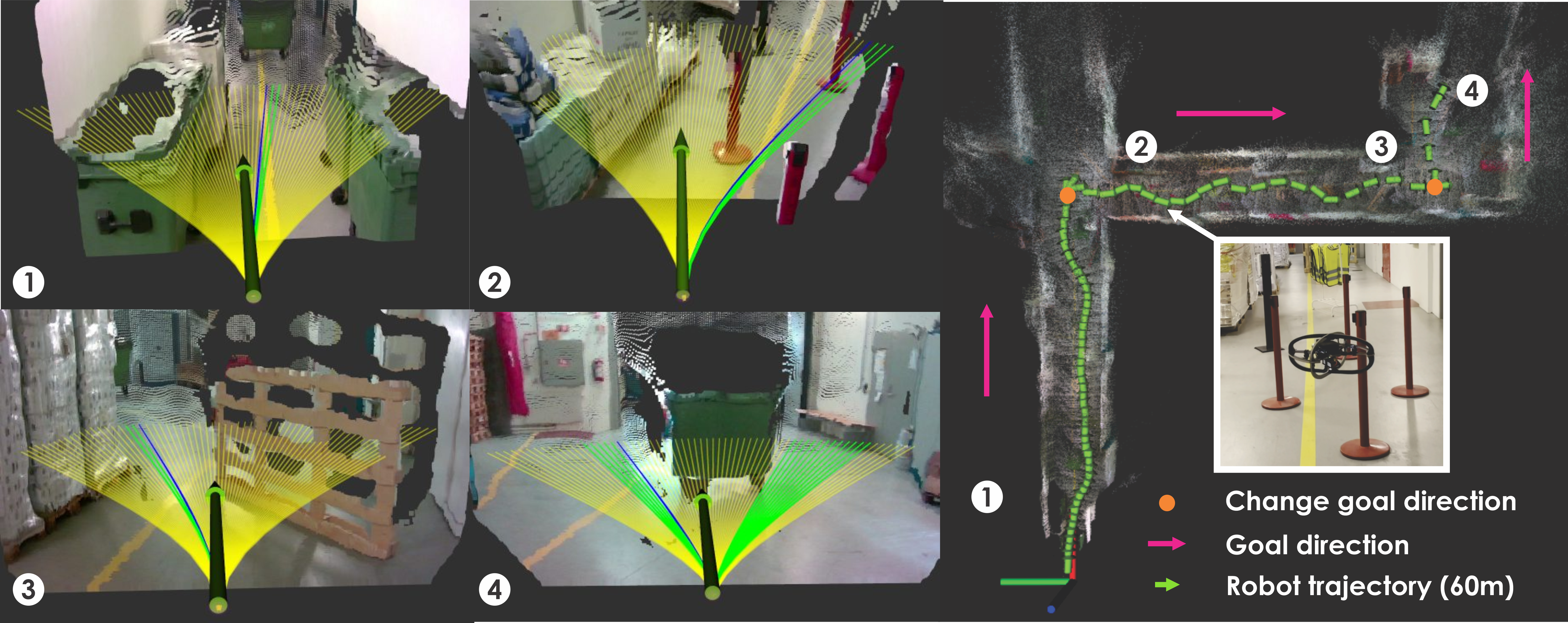}
\vspace{-2ex}
\caption{Experiment in corridors filled with obstacles. The $4$ plots on the left depict specific planning instances where the visualized trajectories are generated based on $s_t$ and the MPL, the green trajectories correspond to the action sequences that pass the collision cost threshold check, while the blue trajectory corresponds to the best action sequence. The right figure presents the reconstructed map of the environment, which is not given to the robot, as well as the goal direction on each segment given by the global planner.} \label{fig:basement_exp}
\vspace{-2ex}
\end{figure*}
%

\subsubsection{System Overview}
LMF inherited the collision-tolerant design of the Resilient Micro Flyer~\cite{Paolo2020RMF}, yet with an increased diameter at $0.43\textrm{m}$ and a mass of $1.2\textrm{kg}$. It integrates a Realsense D455 to obtain depth data, a nadir-facing TFmini $1\textrm{D}$ range sensor, a PixRacer PX4-based autopilot delivering attitude and thrust control, and a Realsense T265 fused with the IMU of the autopilot allowing it to estimate the speed, orientation and angular rate of the robot.
The decision not to assume access to a position estimate reflects our understanding of the challenges aerial robots encounter during fast flight or within perceptually-degraded environments~\cite{Shehryar2020CompSLAM,KTIO_ICRA_2019}. The provided velocity estimates are then used in a fixed-gain velocity controller. Similarly, two Proportional-Integral-Derivative controllers for yaw control and height regulation from the ground using the TFmini sensor are also developed. The ORACLE planner and the low-level controllers are implemented on a Jetson Xavier NX onboard LMF. 


%
\begin{figure}[h!]
\centering
    \includegraphics[width=0.99\columnwidth]{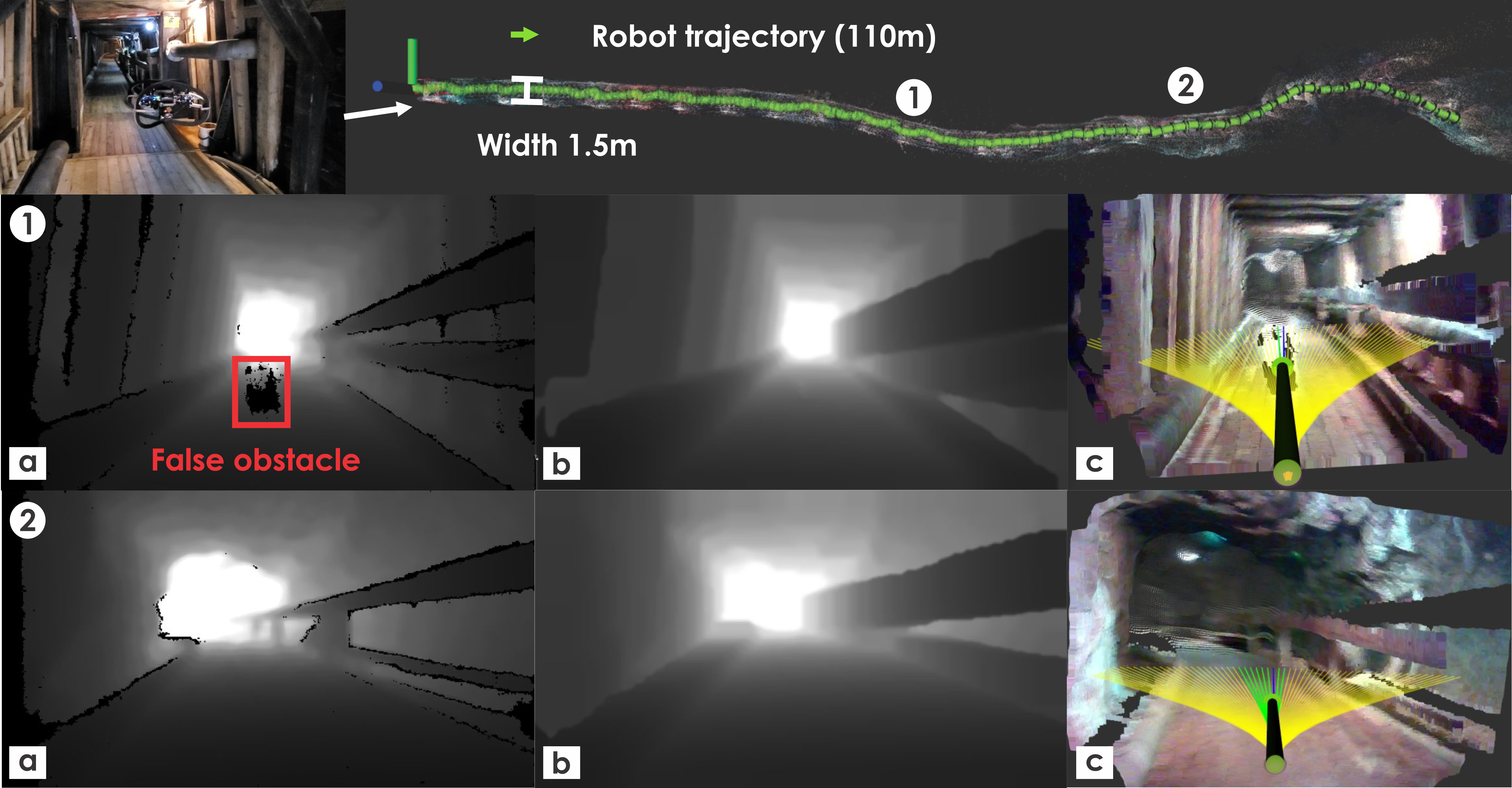}
\vspace{-4ex}
\caption{Field experiment in an underground mine. The top figure illustrates the reconstructed map of the environment, which is not given to the robot. The two bottom rows depict the raw depth image $o_t$ (a), the filtered image $o^\prime_t$ (b), and the visualized trajectories based on $s_t$ and the MPL (c).}\label{fig:lokken_exp}
\vspace{-2ex}
\end{figure}
%

\subsubsection{Experimental Results}

In the first experiment, illustrated in Figure~\ref{fig:basement_exp}, LMF is tasked to follow the goal directions given by a global planner, while navigating safely in a corridor filled with various types of obstacles that were not seen during training. Figure~\ref{fig:basement_exp}.1-4 presents predictions of the CPN at some specific scenarios, where the trajectories are generated only for visualization purposes based on $s_t$ and the MPL using the estimated dynamics models of the velocity and yaw controllers mentioned in subsection~\ref{subsec:implementation}. The green trajectories correspond to the action sequences that pass the collision cost threshold check, while the blue trajectories correspond to the best action sequence. As shown, the visualized trajectories correlate well with the collision cost predicted by the CPN, showing reliable performance of the CPN in various situations.

In the second experiment, LMF traveled a distance of $110$m in a narrow underground environment, which is particularly challenging since the depth image data is not perfect, and the uncertainty in the estimated partial state of the robot is not negligible due to the fairly dark conditions which lead to relatively weak features. Figure~\ref{fig:lokken_exp}.1 demonstrates the effect of the filtering process presented in~\cite{Ku2018InDO}, which reduces the sim-to-real gap and also removes false obstacles. Furthermore, Figure~\ref{fig:lokken_exp}.2 demonstrates the benefit of including in the training environment obstacles with narrow holes that cannot fit the robot, which prevents LMF from erroneously steering to the fence on the right in this situation. At any time, the goal direction is commanded to be the current heading of the robot which in turn allows the system to pick the safe direction that deviates the least from the current heading direction of the robot in this unknown environment.

\subsection{Computational Analysis}
As outlined in Algorithm~\ref{alg:planning_process}, most of the complexity terms of the ORACLE planner come from two operations, namely a) the image filtering process (Line 3) and b) multiple forward passes through the CPN (Lines 9). The inference time of the CPN includes the inference times of the ensemble of CNNs, the Combiner and Prediction networks. Assuming $N_{MC}$ forward passes through the ensemble of CNNs, and $N_{MC}\times N_{\Sigma}$ forward passes through the Combiner network can be run in parallel on the GPU, their complexities are $O(1)$. On the other hand, the inference time of the Prediction network is dominated by the inference time of the LSTM network and has the complexity of $O(H)$ \cite{zhang2016RNN}, assuming that $N_{MC}\times N_{\Sigma} \times N_{MP}$ forward passes through it can be run in parallel on the GPU. 
With $N_{MC}=5, N_{\Sigma}=5, N_{MP}=64, H=18$, on average, the ORACLE loop takes $47.4\textrm{ms}$ in which the inference time through the ensemble of CNNs, the Combiner and Predictor networks, and the filtering time is $6.2\textrm{ms},~2.5\textrm{ms},~8.4\textrm{ms}$, and $20.1\textrm{ms}$, respectively.

\section{CONCLUSIONS}\label{sec:concl}
 In this work, a navigation planner based on a learning-based collision prediction network was proposed and experimentally verified using a custom-designed resilient micro flyer.  
 The algorithm uses a neural network to predict the collision cost for each action sequence in a motion primitives library, and accounts for the uncertainty of the robot's partial state and the neural network model.
 After thresholding the collision cost, a goal direction given by a global planner is used to choose the best action sequence to execute.
 A set of simulation and experimental studies is presented and serves to verify the performance of the proposed approach in fairly cluttered, narrow and visually challenging environments. 

\bibliographystyle{IEEEtran}
\bibliography{./BIB/ORACLE_ICRA_2022}

\end{document}